\documentclass[letterpaper]{article} 
\usepackage{aaai25}  
\usepackage{times}  
\usepackage{helvet}  
\usepackage{courier}  
\usepackage[hyphens]{url}  
\usepackage{graphicx} 
\usepackage{natbib}  
\usepackage{caption} 
\usepackage{algorithm}
\usepackage{algorithmic}
\usepackage{color}
\usepackage{multicol}
\usepackage{multirow}
\usepackage{amsmath}

\usepackage{newfloat}
\usepackage{listings}
\DeclareCaptionStyle{ruled}{labelfont=normalfont,labelsep=colon,strut=off} 
\floatstyle{ruled}
\newfloat{listing}{tb}{lst}{}
\floatname{listing}{Listing}

\title{Enhancing Multi-Hop Fact Verification with Structured \\Knowledge-Augmented Large Language Models}
\author {
    Han Cao\textsuperscript{\rm 1,\rm 2},
    Lingwei Wei\textsuperscript{\rm 1}\thanks{Corresponding author},
    Wei Zhou\textsuperscript{\rm 1},
    Songlin Hu\textsuperscript{\rm 1, \rm 2}
}
\affiliations {
    \textsuperscript{\rm 1}Institute of Information Engineering, Chinese Academy of Sciences, Beijing, China\\
    \textsuperscript{\rm 2}School of Cyber Security, University of Chinese Academy of Sciences, Beijing, China\\
    caohan@iie.ac.cn, weilingwei@iie.ac.cn, zhouwei@iie.ac.cn, husonglin@iie.ac.cn
}

\begin{document}

\maketitle

\begin{abstract}
The rapid development of social platforms exacerbates the dissemination of misinformation, which stimulates the research in fact verification. Recent studies tend to leverage semantic features to solve this problem as a single-hop task. However, the process of verifying a claim requires several pieces of evidence with complicated inner logic and relations to verify the given claim in real-world situations. Recent studies attempt to improve both understanding and reasoning abilities to enhance the performance, but they overlook the crucial relations between entities that benefit models to understand better and facilitate the prediction. To emphasize the significance of relations, we resort to Large Language Models (LLMs) considering their excellent understanding ability. Instead of other methods using LLMs as the predictor, we take them as relation extractors, for they do better in understanding rather than reasoning according to the experimental results. Thus, to solve the challenges above, we propose a novel Structured Knowledge-Augmented LLM-based Network (LLM-SKAN) for multi-hop fact verification. Specifically, we utilize an LLM-driven Knowledge Extractor to capture fine-grained information, including entities and their complicated relations. Besides, we leverage a Knowledge-Augmented Relation Graph Fusion module to interact with each node and learn better claim-evidence representations comprehensively. The experimental results on four common-used datasets demonstrate the effectiveness and superiority of our model.
\end{abstract}

\section{Introduction}

The rapid development of social platforms facilitates the dissemination of misinformation fabricated on purpose. This situation has necessitated the development of a fact verification task, which aims to assess the truthfulness of a given claim with retrieved evidence automatically \cite{DBLP:journals/tacl/GuoSV22, DBLP:journals/llc/ZengAZ21, DBLP:conf/acl/WeiHZYH20, dou2021rumor}. 

Typically, verifying a claim requires several pieces of evidence that exhibit complex inner logic and relations to verify the given claim, which highly demands the capability of multi-step reasoning. Hence, \textbf{multi-hop fact verification} has become an attractive research topic \cite{DBLP:conf/emnlp/ZhuSZZZ023, DBLP:conf/ijcai/OstrowskiAAA21, Zhou19}. 
Unlike traditional verification tasks involving a single inference step, the main challenges of multi-hop verification lie in comprehensively understanding and reasoning complex relations between related evidence pieces. This requires a deep comprehension of the context and a strong reasoning ability for accurate verification.

\begin{table*}[htbp]
\centering 
    {\begin{tabular}{l|p{6cm}|p{8cm}} 
    \hline
    Type & Claim & Evidence  \\ \hline
    One-hop & Little Miss Sunshine was filmed \textcolor{red}{over 30 days}  & Little Miss Sunshine ..., filming began on June and took place \textcolor{red}{over 30 days} in Arizona ...  \\ \hline
    Multi-hop & The \textcolor{red}{Ford Fusion} was introduced for \textcolor{red}{model year 2006}. The Rookie of The Year in the 1997 CART season drives it in the \textcolor{red}{NASCAR Sprint Cup Series}. & \textcolor{red}{Ford Fusion} is \textcolor{blue}{manufactured and marketed by Ford}. Introduced for \textcolor{red}{the 2006 model year}, ..., \textcolor{red}{Patrick Carpentier} competed in the \textcolor{red}{NASCAR Sprint Cup Series}, \textcolor{blue}{driving the} \textcolor{red}{Ford Fusion}, ... \\ \hline
    \end{tabular}}
    \caption{Examples of one-hop and multi-hop claims and evidence. \textcolor{red}{Red marked words} are entities and \textcolor{blue}{blue marked tokens} are inner relations between claim and evidence.} 
    \label{table-0} 
\end{table*}

Existing studies on multi-hop fact verification aim to improve both understanding and reasoning abilities \cite{DBLP:conf/aaai/ZhangZZ24, DBLP:conf/aaai/SiZZ23, DBLP:conf/emnlp/PanLKN23}. \citet{DBLP:conf/emnlp/PanLKN23} leverage a question-answer framework to extract relevant evidence as comprehensively as possible and enhance the model understanding capacity. 
To promote the reasoning ability, \citet{DBLP:conf/aaai/ZhangZZ24} and \citet{DBLP:conf/aaai/SiZZ23} construct graph structure and utilize graph fusion methods to model complex relations between coarse-grained evidence debunks. However, all nodes are treated equally and construct fully connected claim graphs, overlooking the inner relations between claims, entities, and each piece of evidence. 

Despite recent advancements in multi-hop fact verification, these methods often struggle with handling the complex relationships between fine-grained knowledge for the following two challenges.
1) 
The multi-source evidence pieces and some informality or ambiguity in language aggregate the insufficient model understanding. According to \citet{DBLP:conf/aaai/ZhangZZ24}, recent models tend to learn a shortcut to predict labels, instead of fully understanding the inner logic and contextual relations. This hinders the accurate understanding of context. 
Although some preliminary works leverage Large Language Models (LLMs) to predict factual verdicts through prompt tuning \cite{DBLP:conf/ijcnlp/ZhangG23, DBLP:conf/www/ChoiF24}, they usually achieve unsatisfactory performance compared to other methods based on small models, especially in multi-hop fact verification.  
2) These methods overlook the complicated relations between fine-grained knowledge such as entities in evidence, which are crucial for understanding the multi-step dependencies necessary for verification. 
As shown in Table~\ref{table-0}, a one-hop claim only needs verifying the very aspect, whereas a multi-hop claim requires multi-step thinking, e.g. the relation between \textit{Ford Fusion} and \textit{NASCAR Sprint Cup Series}, which is crucial to capture the logic and help model better understand it accurately.  
These relations implicitly reveal the aspects that need to be verified, enabling the model to adapt to varying contexts and learn from diverse evidence types.
Therefore, how to effectively improve disambiguate contextual semantics and capture complex relationships between fine-grained knowledge in evidence remains a challenging problem for multi-hop veracity verification.

To alleviate the above challenges, this paper investigates the use of large language models (LLMs) for fine-grained knowledge extraction and their integration with graph networks to enhance reasoning in multi-hop fact verification tasks.
First, since LLMs are capable of understanding context and holding vast amounts of knowledge, we leverage LLMs to extract fine-grained knowledge in both claim and evidence, {such as \textit{Ford Fusion} and \textit{2006 model year}}. Besides, we fine-tune LLMs with structured knowledge to fully adapt the model to extract and utilize specific types of structured knowledge.
We further fuse the extracted fine-grained knowledge with graph-based networks to enhance the reasoning ability, which can model the complex dependencies among knowledge triples for reasoning. 
In this way, we can better understand and reason multi-hop connections and relational information for better verification. 

In this paper, we propose a novel Structured Knowledge-Augmented LLM-based Network (LLM-SKAN) to fully understand and reason with the augmentation of fine-grained knowledge for multi-hop fact verification.
Specifically, we enhance LLM with structured knowledge like triplets to extract complicated fine-grained relations between entities to unfold the inner logic for a better understanding of the model. 
Now that LLM is poor in reasoning, we only leverage its understanding capacity and design a reasoning module with small-scale models. We construct heterogeneous graphs using augmented data and utilize a small-scale Graph Neural Network (GNN) to learn comprehensive representations and make predictions. LLM-SKAN incorporates a small model with LLMs, taking advantage of the powerful capability of LLMs for NLP tasks and the trainable and excellent prediction performance of small models.

We conduct experiments on several common-used multi-hop fact verification datasets FEVER \cite{DBLP:conf/emnlp/Thorne0C0M18}, and HOVER \cite{DBLP:conf/emnlp/JiangBZD0B20} to assess the effectiveness of LLM-SKAN. The experimental results show the effectiveness and superiority of our proposed model. The extensive experiments demonstrate that the extractor can obtain relations facilitating the model to make correct predictions.

Our main contributions are as follows:
\begin{itemize}
    \item We proposed a novel Structured Knowledge-Augmented LLM-based Network for multi-hop fact verification, incorporating a large language model to enhance the model with structured knowledge and improve its understanding of inner logic\footnote{The code will be released in https://github.com/HanCao12/LLM-SKAN}.
    \item We proposed a novel LLM-based entity triplet extraction module to emphasize and capture complicated fine-grained information between claims, each piece of evidence, and entities, to unfold the inner logic that contributes to better multi-step thinking. 
    \item To evaluate the performance of our proposed method, we carry out experiments on 4 commonly used fact verification datasets. Our model outperforms the comparison methods, which demonstrates the effectiveness and superiority of the proposed model.
\end{itemize}

\section{Related Work}
Fact verification aims to predict the verdicts of check-worthy claims with several retrieved evidence. Traditional fact verification approaches only utilize textual information to make predictions \cite{DBLP:conf/aaai/ZhangZZ24, DBLP:conf/acl/KimPKJTC23, deberta}, which fails to deal with claims that need multi-hop consideration. Hence, multi-hop fact verification has become a research hotspot. Besides, LLMs have made significant developments and have been applied to fact verification tasks. In this section, we will report on the related work in these three research fields.

\begin{figure*}[t]
  \begin{center}
  \includegraphics[width=0.95\linewidth]{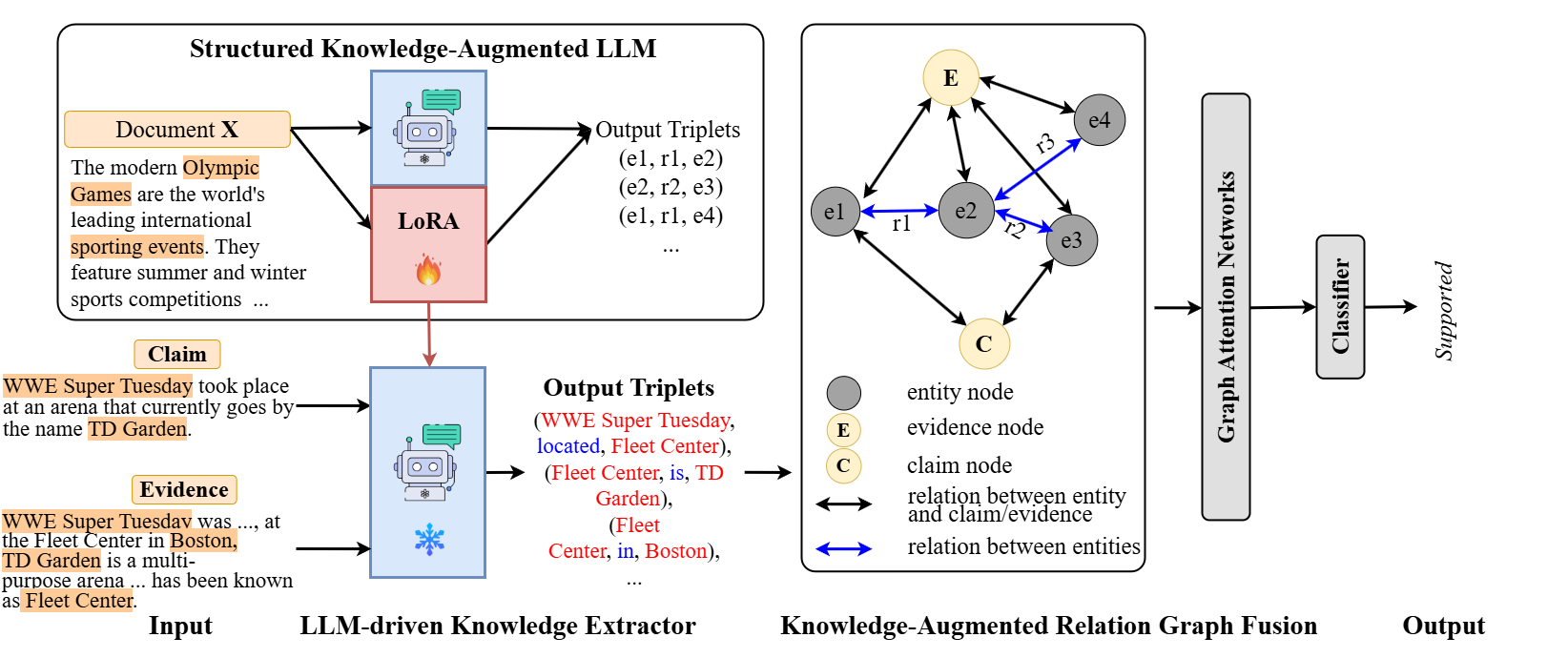}
  \end{center}
  \caption{The framework of our proposed LLM-SKAN. This framework consists of four main components: (1) Structured knowledge-augmented LLM, aiming to fine-tune the LLM to capture more accurate fine-grained relations; (2) LLM-driven knowledge extractor, aiming to extract fine-grained knowledge; (3) Knowledge-augmented relation graph fusion, aiming to learn comprehensive representations through augmented graph; (4) Fact verification, utilizing the comprehensive representation to predict the label.}
  \label{f:1}
\end{figure*}

\subsection{Fact Verification}
Research on unimodal fact verification typically involves verifying text-only claims using textual evidence, such as metadata of the claim, documents retrieved from knowledge bases, or tabular evidence \cite{Wang17, Aly21, DBLP:journals/nlpj/PanchendrarajanZ24, DBLP:conf/aaai/Gong0W0024}. 
\citet{Wang17} incorporated additional metadata, such as the speaker's profile, to verify claims using a Convolutional Neural Network (CNN).
\citet{Li21} proposed a multi-task learning method that integrates data features with paragraph-level evidence for scientific claim verification. \citet{Chen21} and \citet{Zhou19} utilized entities extracted from textual contents to construct entity graphs, attempting to learn more granular data representations. Some researchers have tried to leverage structured resources, such as tabular data, to pursue better performance. For example, \citet{Gu22} serialized table evidence as sequential data and concatenated it with the claim to assess its verdict. \citet{Wangfei21} learnt the salient semantic representations for fact verification to deal with the unbalanced vocabulary of statements and evidence. \citet{DBLP:conf/aaai/Gong0W0024} leveraged a heterogeneous graph to fuse structured and unstructured data.
These approaches leverage various claim-evidence interaction methods to deal with text-only fact verification and demonstrate satisfactory performance on unimodal fact verification.

\subsection{Multi-hop Fact Verification}
Multi-hop fact verification aims to detect claims that need multi-step thinking, for there are complicated inner relations between entities, claims, and evidence \cite{DBLP:conf/emnlp/JiangBZD0B20, DBLP:conf/ijcai/OstrowskiAAA21, DBLP:conf/aaai/ZhangZZ24}.
\citet{DBLP:conf/emnlp/JiangBZD0B20} noticed that common-used datasets for fact verification only focus on single-hop tasks and lack multi-hop claims in these datasets. Hence, they proposed a new dataset targeted at multi-hop fact verification. \citet{DBLP:conf/emnlp/PanLKN23} utilized a question-answering framework to retrieve several pieces of relevant evidence to make predictions. \citet{DBLP:conf/aaai/SiZZ23} leveraged a graph-based framework to learn word-salience representations and fuse claim and evidence information to solve multi-hop tasks. \citet{DBLP:conf/aaai/ZhangZZ24} investigated the shortcut path problem and proposed a causal intervention method to eliminate it.
These methods emphasize the significance of multi-hop logic acquisition to effectively deal with multi-hop fact verification.

\subsection{Large Language Models}
In recent years, Large Language Models (LLMs) have developed rapidly and have a significant performance in NLP tasks \cite{DBLP:conf/nips/BrownMRSKDNSSAA20, openai2024gpt4technicalreport, DBLP:journals/corr/abs-2307-09288}. 
GPT-2 was proposed to deal with language tasks with large-scale parameters \cite{radford2019language}. With the increment of scale of pre-trained data and parameters, GPT-3 \cite{DBLP:conf/nips/BrownMRSKDNSSAA20} held a new era for LLMs and has excellent performance in many research fields. Inspired by GPT-3, LLaMa \cite{DBLP:journals/corr/abs-2302-13971}, Mistral \cite{DBLP:journals/corr/abs-2310-06825}, and Vicuna \cite{vicuna} were proposed successively based on the pre-training corpus and training methods of GPT-3. Llama2 made a great improvement in understanding and reasoning abilities based on LLaMa \cite{DBLP:journals/corr/abs-2307-09288}. To further improve the capability of LLMs, GPT-4 \cite{openai2024gpt4technicalreport} was proposed and pre-trained by multimodal data, enabling LLMs to deal with multimodal tasks.
{Nevertheless, some works have proven the significance of LLMs in relation extraction tasks \cite{DBLP:conf/acl/WadhwaAW23, DBLP:conf/emnlp/WanCMLSLK23}, LLMs' reasoning capability in fact verification is limited for they may forget some contextual information in reasoning when they are dealing with long documents. This makes them fail to get better performance in multi-hop tasks. Thus, we only utilize LLMs to extract fine-grained knowledge rather than to predict verdicts directly.}

\section{Methodology}
In this section, we present the Structured Knowledge-Augmented LLM-based Network (LLM-SKAN) in detail for multi-hop fact verification. We begin by defining the task, after which we introduce the overall framework of LLM-SKAN. After, we'll go over the details of the proposed method.

\subsection{Task Definition}
Multi-hop fact verification aims to verify the truthfulness of a given claim with several pieces of retrieved evidence that have complicated inner relations. Let $\mathcal{D} = \{C, E_1, E_2, ... , E_m\}$ be the claim-evidence pair of the dataset, where $C$ denotes the claim and $E_1, E_2, ... , E_m$ denotes its relevant evidence. Each pair has a verdict $y\in \mathcal{Y}$. The goal is to find a function $F: \mathcal{D} \rightarrow \mathcal{Y}$ that maps the data to the label set and makes predictions.

\subsection{Overall Architecture}
Our objective is to extract fine-grained entities and capture complex relations to verify the claim's truthfulness. Hence, we propose a novel \textit{Structured Knowledge-Augmented LLM-based Network} for multi-hop fact verification. Fig. \ref{f:1} illustrates the overall architecture of LLM-SKAN, which mainly consists of the following components:
\begin{itemize}
    \item [$\bullet$] \textbf{Structured knowledge-augmented LLM:} We fine-tune it with a relation extraction task to fit the LLM with our objective.
    \item [$\bullet$] \textbf{LLM-driven Knowledge Extractor:} To highlight the importance of fine-grained entities and complicated relations, we leverage an LLM-driven Knowledge Extractor to extract structured entity triplets.
    \item [$\bullet$] \textbf{Knowledge-Augmented Relation Graph Fusion:} To learn better representations, we design a heterogeneous graph fusion module to capture comprehensive claim-evidence relations and interactions.
    \item [$\bullet$] \textbf{Fact Verification:} We integrate the fused claim-evidence representation as the input to the classifier to predict the label of each claim-evidence pair.
\end{itemize}

\subsection{Structured Knowledge-Augmented LLM}
We fine-tune Llama2 to fully explore the capability of LLM instead of directly utilizing the knowledge of LLM learned in pre-training. In detail, following \citet{DBLP:conf/iclr/HuSWALWWC22}, we deliberately choose a document-level entity relation extraction dataset DocRED-FE \cite{DBLP:conf/icassp/WangXSZXL23} and utilize the Low-Rank Adaptation (LoRA) mechanism to fine-tune the LLM to extract ErE triplet. 

The loss of fine-tuning is calculated by:
\begin{equation}
\label{eq-13}
    \mathcal{L}_{ft} = -\sum_{i=1}^{|\mathcal{D}|} l_ilog(\hat{l}_i),
\end{equation}
where $l_i$ and $\hat{l_i}$ denote the predicted label and true label of extracted triplet respectively. Specifically, if the extracted triplet is in the ground truth, $\hat{l_i}=1$, otherwise $\hat{l_i}=0$.

\subsection{LLM-driven Knowledge Extractor}
Noticing the crucial impact of fine-grained information like entities and structured data compared to unstructured data, we design an LLM-driven Knowledge Extractor. It aims to extract relevant entities of claim and evidence and capture entity-relation-entity (ErE) triplets to obtain structured data. Different from knowledge-graph-based methods, there is no need for extra interfaces and a shortened search time. With the fine-tuned LLM, we obtain triplet set $\mathcal{T} = \{(Ent_h, R, Ent_t)\}^{|\mathcal{T}|}$ through the following prompt:
\begin{quote}
\textit{Please extract entities in the given text and the relations between entities. Let's think step by step. Please return in this form: (entity, relation, entity). Here is the text: [TEXT].}
\end{quote}
where $Ent_h, Ent_t$ denote entities and $R$ denotes relation. 

\subsection{Knowledge-Augmented Relation Graph Fusion}
Several methods demonstrate the effectiveness of structured data, like graphs, dealing with fact verification tasks, compared to unstructured data \cite{DBLP:conf/aaai/SiZZ23, cao2024multisourceknowledgeenhancedgraph, DBLP:conf/coling/WangLHD22}. Therefore, we propose a Knowledge-Augmented Relation Graph Fusion module to integrate claims, evidence and entities more comprehensively.

\paragraph{Knowledge-Augmented Relation Graph Construction} 
Based on the claim, evidence, and extracted triplet $\mathcal{T}$, we construct a relation graph to promote a better combination of coarse-grained and fine-grained information. For each sample $\mathcal{D}$, we define an undirected graph $G=\{V, E\}$, where $V$ and $E$ refer to node and edge sets.

\textbf{Nodes.} First, we add each claim and each piece of evidence into the node set. We set entities extracted from them as nodes and integrate duplicated entities into one node. We utilize a pre-trained model DeBERTa \cite{deberta} mean-pool the hidden state of the last layer of each token to extract node features $c$, $e$, and $ent$. Thus, we get the node set $V = \{c, e, ent\}$.

\textbf{Edge.} We first link entities to claims and evidence according to where they are extracted. Besides, each two entities is linked if there is an ErE triplet. For edge features, we manually set the relation "belong to" to each claim- and evidence-entity edge to avoid inconsistency. Thus, each edge has a relation, and we utilize the same pre-trained model to extract edge features $r$. Then, we get the edge set $E = \{r\}$.

\paragraph{Knowledge-Augmented Relation Graph Representation Learning}
After the graph construction, we leverage a graph neural network to learn the graph representations to fully integrate graph features.
During the graph fusion process, each node embedding $v \in V$ is iteratively updated to capture complex interactions between the claim, evidence, and entities linked by relations. The equation for the $l$-th layer is given by:
\begin{eqnarray}\label{eq-6}
    v_i^{(l)} = \gamma_{i,i}\Theta v_i^{(l-1)} + \sum_{j\in \mathcal{N}(i)}\gamma_{i,j}\Theta v_j^{(l-1)},
\end{eqnarray}
where $\Theta$ denotes the transformation parameters and $\gamma_{i,j}$ is the attention score between node $i$ and its neighbor node $j$:
\begin{eqnarray}\label{eq-7}
    \gamma_{i,j} = \frac{\text{exp}(a^T\sigma(\Theta[v_i||v_j||r_{i, j}]))}{\sum_{k\in \mathcal{N}(i)\cup \{i\}}\text{exp}(a^T\sigma(\Theta[v_i||v_k||r_{i, k}]))},
\end{eqnarray}
where $||$ is the concatenation operation, $\sigma$ stands for Leaky Rectified Linear Unit.
After the graph fusion, we take the claim node feature of each relation graph as the graph representation $\tilde{v}$.

Extracted structured knowledge unfolds the inner logic between claim and evidence explicitly and graph fusion facilitates the model to learn comprehensive and representative fused features with the augmented data.

\subsection{Fact Verification}
We use the fused representation $\tilde{v}$ as input to the category classifier to predict the label, which consists of a 2-layered fully connected network. The prediction process is carried out as follows:
\begin{equation}
\label{eq-12}
    \hat{y} = softmax(W^1\sigma(W^0\tilde{v})),
\end{equation}
where $W^0$ and $W^1$ are learnable parameters and $\hat{y}$ is the predicted label.

In the training stage, we utilize the cross-entropy loss as the training loss:
\begin{equation}
\label{eq-14}
    \mathcal{L}_{cls} = -\sum_{i=1}^{|C|} y_ilog(\hat{y}_i),
\end{equation}

\begin{table}
\centering 
    \begin{tabular}{l|ccc} 
    \hline
    Dataset & Train & Dev & Test \\ \hline
    FEVER & 145,449 & 19,998 & 19,998 \\ 
    2-hop HOVER & 9,052 & 1,126 & 1,333 \\
    3-hop HOVER & 6,084 & 1,835 & 1,333\\
    4-hop HOVER & 3,035 & 1,039 & 1,333\\ \hline
    \end{tabular}
    \caption{The statistic of FEVER and HOVER datasets.} 
    \label{table-1} 
\end{table}

\begin{table*}[htbp]
\centering 
    {\begin{tabular}{l|cc|cc|cc|cc} 
    \hline
   \multirow{2}{*}{\centering{\textbf{Model }}} & \multicolumn{2}{c|}{\textbf{FEVER}} &\multicolumn{2}{c|}{\textbf{2-hop HOVER}} &\multicolumn{2}{c|}{\textbf{3-hop HOVER}} &\multicolumn{2}{c}{\textbf{4-hop HOVER}}\\ \cline{2-9}
   & Acc & FEVER & Acc & FEVER & Acc & FEVER & Acc & FEVER \\ \hline
   DeBERTa \cite{deberta} & 65.37 & 61.81 & 72.94 & 68.88 & 71.67 & 67.98 & 70.34 & 67.12 \\
   GEAR \cite{Zhou19} & 71.60 & 67.10 & 73.50 & 69.17 & 72.33 & 69.08 & 71.79 & 67.99 \\
   EvidenceNet \cite{10.1145/3485447.3512135} & 73.31 & 69.40 & 73.95 & 69.89 & 73.23 & 68.50 & 72.46 & 68.93 \\
   CO-GAT \cite{DBLP:journals/corr/abs-2405-10481} & 77.27 & 73.59 & 77.85 & 73.51 & 76.40 & 73.06 & 75.11 & 71.97 \\
   SaGP \cite{DBLP:conf/aaai/SiZZ23} & \underline{78.47} & \underline{74.52} & \underline{77.90} & \underline{73.84} & \underline{76.78} & \underline{73.22} & \underline{76.01} & {72.66} \\\hline
   HiSS \cite{DBLP:conf/ijcnlp/ZhangG23} & 62.30 & 59.38 & 70.41 & 67.94 & 68.26 & 64.17 & 67.33 & 63.20\\
   MultiKE-GAT \cite{cao2024multisourceknowledgeenhancedgraph} & - & - & 77.04 & 73.79 & 76.28 & 73.10 & 75.85 & \underline{72.73} \\
   \textbf{LLM-SKAN (ours)} & \textbf{79.25} & \textbf{75.32} & \textbf{79.90} & \textbf{75.20} & \textbf{78.23} & \textbf{74.09} & \textbf{77.95} & \textbf{73.78} \\ 
   \hline
    \end{tabular}}
    \caption{Result of fact verification tasks on 4 datasets. We use the FEVER score (\%) and Accuracy (Acc, \%) to evaluate the performance. \textbf{Bold} denotes the best performance. \underline{Underline} denotes the second-best performance.} 
    \label{table-2} 
\end{table*}

\section{Experimental Setups}

\subsection{Datasets}
To evaluate the effectiveness of LLM-SKAN for both single-hop and multi-hop fact verification tasks, we choose 4 public benchmarks, FEVER \cite{DBLP:conf/emnlp/Thorne0C0M18} and 2-, 3-, and 4-hop HOVER \cite{DBLP:conf/emnlp/JiangBZD0B20}, to conduct experiments. 
\textbf{FEVER} collects more than 180,000 human-generated claims with retrieved evidence from Wikipedia\footnote{https://www.wikipedia.org/} for single-hop fact verification and they are categorized into \textit{Supported}, \textit{Refuted}, and \textit{NEI}.
\textbf{HOVER} contains more than 15,000 claims that need multi-hop thinking to verify and retrieves evidence from Wikipedia for multi-hop fact verification. Each claim is labelled \textit{Supported} and \textit{Not-Supported}. It contains three sub-datasets, 2-hop, 3-hop, and 4-hop HOVER. The statistics are shown in Table~\ref{table-1}.

\subsection{Baselines}

We compare it to several fact verification approaches to assess the performance of LLM-SKAN.
\textbf{DeBERTa} \cite{deberta} leverages the pre-trained model DeBERTa to extract textual features to make predictions. 
\textbf{GEAR} \cite{Zhou19} leverages a graph neural network to fuse claim and evidence features and make predictions.
\textbf{EvidenceNet} \cite{10.1145/3485447.3512135} selects relevant and useful sentences from document-level evidence and uses a gating mechanism and symmetrical interaction attention mechanism to predict the label.
\textbf{CO-GAT} \cite{DBLP:journals/corr/abs-2405-10481} uses multiple sentence-level evidence and a GAT-based method to verify claims.
\textbf{SaGP} \cite{DBLP:conf/aaai/SiZZ23} leverages perturbed graph neural network and selects rational subgraphs to make predictions and give explanations.
\textbf{HiSS} \cite{DBLP:conf/ijcnlp/ZhangG23} leverages LLMs and prompt-tuning mechanism to decompose the claim and verify each subclaim to make final prediction.
\textbf{MultiKE-GAT} \cite{cao2024multisourceknowledgeenhancedgraph} utilizes LLMs to extract fine-grained entities without relations to make predictions.

\begin{table*}[htbp]
\centering 
    {\begin{tabular}{l|cc|cc|cc|cc} 
    \hline
   \multirow{2}{*}{\centering{\textbf{Model }}} & \multicolumn{2}{c|}{\textbf{FEVER}} &\multicolumn{2}{c|}{\textbf{2-hop HOVER}} &\multicolumn{2}{c|}{\textbf{3-hop HOVER}} &\multicolumn{2}{c}{\textbf{4-hop HOVER}}\\ \cline{2-9}
    & Acc & FEVER & Acc & FEVER & Acc & FEVER & Acc & FEVER \\ \hline
    \textbf{LLM-SKAN} & \textbf{79.25} & \textbf{75.32} & \textbf{79.90} & \textbf{75.20} & \textbf{78.23} & \textbf{74.09} & \textbf{77.95} & \textbf{73.78} \\ 
    \ \ \ -w/o KnoE & 78.08 & 73.24 & 78.05 & 74.66 & 77.85 & 73.28 & 76.57 & 72.69 \\ 
    \ \ \ -w/o RGC & 77.04 & 72.28 & 77.17 & 73.42 & 76.80 & 72.30 & 75.92 & 72.22 \\
    \ \ \ -w/o RGF & 77.40 & 72.80 & 77.44 & 73.60 & 76.97 & 72.80 & 76.37 & 72.75 \\\hline
    \end{tabular}}
    \caption{Results of ablation study on 4 datasets. We use the FEVER score (\%) and Accuracy (Acc, \%) to evaluate the performance. \textbf{Bold} denotes the best performance. KnoE, RGC, and RGF denote the knowledge extractor, the relation graph construction, and the relation graph fusion respectively} 
    \label{table-20} 
\end{table*}

\subsection{Implementation Details}

We use a Tesla V100-PCIE GPU with 32GB memory for all experiments and implement our model via the Pytorch framework. 
The number of attention heads is set to 8. The batch size is 24. We set the learning rate as 2e-4. To keep consistency, we set the number of nodes of each relation graph to the maximum 20. If the origin graph has fewer nodes, we manually add isolated nodes. We fine-tune Llama2-7b \cite{DBLP:journals/corr/abs-2307-09288} as the extractor.

\subsection{Evaluation Metrics}
Following \citet{thorne-etal-2018-fever}, we utilize Accuracy and the FEVER score as the evaluation metrics. FEVER score takes evidence selection into account, which can reflect the accuracy of both the prediction and the evidence selection. 

\section{Results and Discussion}
\subsection{Overall Performance}

We conduct the experiments on four datasets and the experimental results are shown in Table \ref{table-2}. For the multi-hop fact verification, LLM-SKAN outperforms other approaches that are designed for single-hop tasks purposely. Compared to non-graph-based method DeBERTa \cite{deberta}
, graph-based approaches perform much better, elucidating that structured data can capture more correlations than unstructured sequential data. Besides, LLM-SKAN outperforms other graph-based approaches, demonstrating that entity-relation-entity triplets enhance the performance and effectiveness of graph fusion. Compared to the LLM-based method, our proposed LLM-SKAN significantly improves the performance, which indicates that few-shot fine-tuning can make LLMs more powerful in solving multi-hop fact verification.

Furthermore, we conduct experiments on single-hop fact verification. LLM-SKAN obtains competitive and satisfactory performance compared to other methods, which demonstrates that the fine-grained information and relations between entities are helpful to the single-hop tasks as well.

Overall, the experimental results demonstrate that LLM-SKAN has the outstanding capability of handling multi-hop fact verification tasks through fine-grained entities and complicated relations between entities. It also shows that LLM-SKAN can solve single-hop tasks excellently and perform at a comparable level.

\subsection{Ablation Study}

We conduct the ablation study to analyze key components of LLM-SKAN. We remove each component including the knowledge extractor (KnoE), the relation graph construction (RGC), and the relation graph fusion (RGF), respectively. The ablation results are shown in Table~\ref{table-20}. Under both single-hop and multi-hop fact verification tasks, the full model achieves the best performance on all datasets consistently. 

Specifically, we first remove the LLM-driven Knowledge Extractor, and the results decline dramatically, especially in multi-hop tasks. This indicates that fine-grained information like entities is crucial to fact verification tasks, and ErE triplets capture complex correlations important to multi-hop thinking. 
Besides, we investigate the effectiveness of relation graphs. Firstly, we replace the relation graphs with fully connected graphs and remove the edge features of relations. It can be observed that relation graph construction plays an essential role in both single-hop and multi-hop tasks, and triplets cannot work well and enhance performances. Then, instead of relation graph fusion, we utilize a sequential attention mechanism to fuse claim-evidence representations. The performance degrades obviously, indicating that sequential attention mechanisms cannot capture complicated semantic information and learn better representations.

Overall, these results give an insightful investigation of the efficacy of each component of LLM-SKAN and demonstrate the effectiveness and superiority of our model LLM-SKAN.

\begin{figure}[t]
  \begin{center}
  \includegraphics[width=0.85\linewidth]{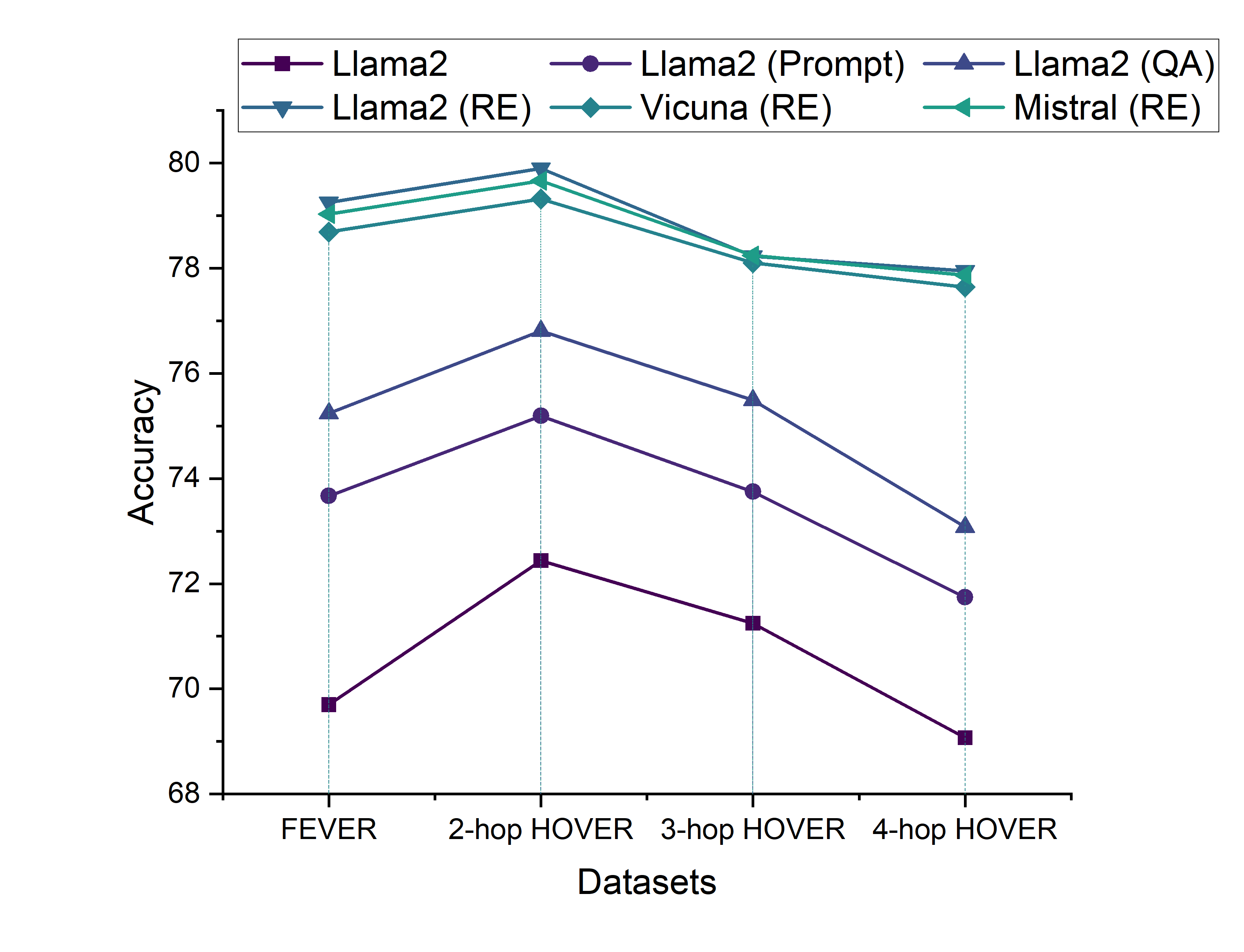}
  \end{center}
  \caption{The comparison of different relation extraction methods. We compare Llama2 to Mistral-7B and Vicuna-7B. Prompt, QA and RE denote the prompt-tuning, fine-tuning based on the question-answering task, and fine-tuning based on the relation extraction task, respectively.}
  \label{f:2}
\end{figure}

\begin{figure}[t]
  \begin{center}
  \includegraphics[width=0.7\linewidth]{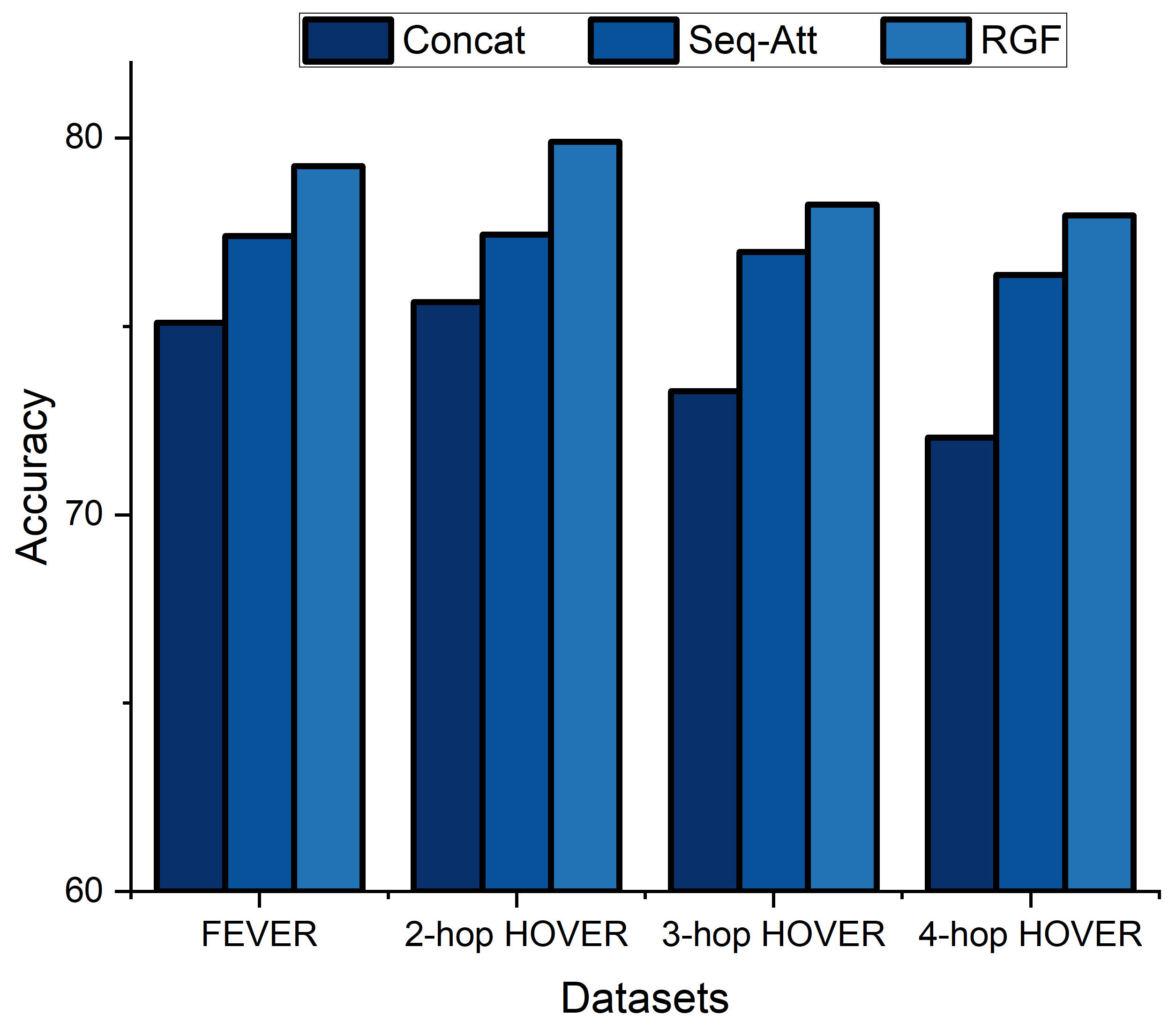}
  \end{center}
  \caption{The comparison of different claim-evidence fusion methods. Concat, Seq-Att, and RGF denote using the simple concatenation, the sequential attention mechanism, and the relation graph fusion module to model complex relations for verification, respectively.}
  \label{f:3}
\end{figure}

\begin{table*}[htbp]
\centering 
    {\begin{tabular}{p{6cm}|p{8cm}|c} 
    \hline
    Claim & Evidence & Relation graph \\ \hline
    The \textcolor{red}{park (e1)} at which \textcolor{red}{Tivolis Koncertsal(e2)} is \textcolor{blue}{located (r1)} \textcolor{blue}{opened (r2)} on \textcolor{red}{15 August 1843 (e3)}. & \textcolor{red}{Tivolis Koncertsal(e2)} is an amusement park and pleasure garden in Copenhagen, Denmark. The \textcolor{red}{park (e1)} \textcolor{blue}{opened (r2)} on \textcolor{red}{15 August 1843 (e3)} ... & \begin{minipage}[b]{0.15\columnwidth}
		\raisebox{-.8\height}{\includegraphics[width=\linewidth]{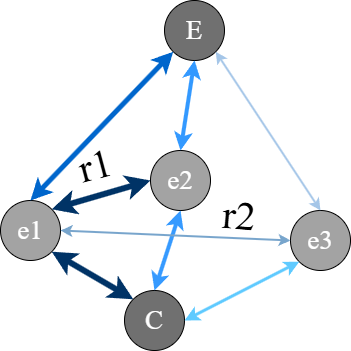}}
	\end{minipage} \\ \hline
    The \textcolor{red}{Ford Fusion (e1)} \textcolor{blue}{was introduced (r1)} for model year \textcolor{red}{2006 (e2)}. \textcolor{red}{The Rookie of The Year (e3)} in the 1997 CART season \textcolor{blue}{drives (r2)} it in the \textcolor{red}{NASCAR Sprint Cup Series (e4)}. & \textcolor{red}{Ford Fusion (e1)} is {manufactured and marketed by Ford}. Introduced for the 2006 model year, ..., \textcolor{red}{Patrick Carpentier (e5)} \textcolor{blue}{competed in (r3)} the \textcolor{red}{NASCAR Sprint Cup Series (e4)}, \textcolor{blue}{driving the (r2)} \textcolor{red}{Ford Fusion (e1)}, ... & \begin{minipage}[b]{0.18\columnwidth}
		\raisebox{-1\height}{\includegraphics[width=\linewidth]{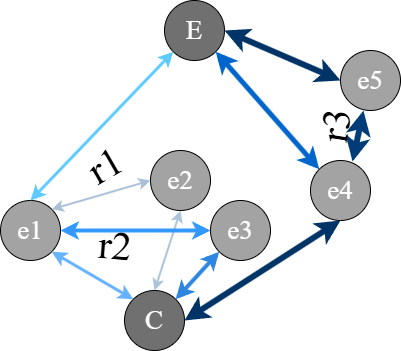}}
	\end{minipage} \\ \hline
    \end{tabular}}
    \caption{Some correctly classified examples by LLM-SKAN. \textcolor{red}{Red marked words} are entities and \textcolor{blue}{blue marked tokens} are inner relations between claim and evidence. The thickness of the edges demonstrates the importance.} 
    \label{table-4} 
\end{table*}

\subsection{Module Analysis}
We conduct several experiments to further demonstrate the effectiveness of LLM-SKAN. Specifically, we compare our knowledge extraction to several methods and investigate the validity of different fusion approaches.

\subsubsection{Impact of Knowledge Extractor}
We first replace the LLM-driven Knowledge Extractor with several LLM-based methods. The results are shown in Fig~\ref{f:2}. 
It demonstrates that LLMs without tuning can perform well, but there is still a huge gap in the performance in fact verification tasks, illustrating that fine-tuning can unearth the potential capability of LLMs. Furthermore, parameter fine-tuned models outperform the model fine-tuned through prompts, which elucidates that in-context learning is limited to enhancing the capacity of LLMs. 
Besides, we compare the performance of Llama2-7B with Mistral-7B and Vicuna-7B. The discrepancy is relatively small but Llama2-7B still outperforms other LLMs, further indicating the reason why we choose Llama2-7B.

Thus, our proposed triplet extractor has the best performance and the results further demonstrate the effectiveness of the extractor fine-tuned based on relation extraction tasks.

\subsubsection{Impact of Fusion Methods}
We compared it to several fusion methods to illustrate the capability of our fusion method. The results are shown in Fig~\ref{f:3}.
The model based on relation graph fusion outperforms the other two models, further indicating that structured data with structured fusion methods can capture the complicated relations between claim and evidence, and perform thorough interactions to learn comprehensive representations. Besides, the model based on the sequential attention mechanism performs better than the model based on concatenation, showing that the sequential mechanism can capture some relations to some extent.

Therefore, this experiment further demonstrates the efficacy of our proposed fusion method.

\subsection{Case Study}
We analyze representative examples that are correctly classified by our model. They are shown in Table~\ref{table-4}\footnote{To keep conciseness, we only choose one piece of evidence to show the efficacy of our model.}. 
The LLM-driven Knowledge Extractor can successfully extract fine-grained entities and their complicated relations. For example, the first claim contains three main entities that contribute to the prediction, and through our model, these three key entities are exactly captured, together with their relations.
Based on these triplets, the relation graphs are constructed shown in the third column in Table~\ref{table-4}. These relation graphs concisely demonstrate the relations between claims, evidence, and entities, which facilitate the fusion module to easily and comprehensively pass the useful information and learn comprehensive representations.

Hence, our model LLM-SKAN is capable of capturing complex relations and fine-grained information and constructing concise and informative relation graphs for classification.

\section{Conclusion}
This work has investigated the multi-hop fact verification tasks that need multi-step thinking and several pieces of evidence to predict the verdict of a given claim. 
{We propose a novel Structured Knowledge-Augmented LLM-based Network (LLM-SKAN) to enhance the model's capability of understanding complicated inner logic and accurate reasoning. }
Specifically, we first design a novel fine-tuned LLM-driven Knowledge Extractor, aiming to capture fine-grained information and extract complicated relations between claims, evidence, and entities.
Moreover, we propose Knowledge-Augmented Relation Graph Fusion to fully leverage the structured data and construct concise and informative relation graphs to thoroughly interact with each node and learn comprehensive representations. 
The experimental results on four commonly used datasets show that LLM-SKAN has been proven to be capable of effectively dealing with both single-hop and multi-hop fact verification tasks in comparison with other competitive methods. 

\appendix
\section{Acknowledgments}
This work was supported by the National Key Research and Development Program of China (No. 2022YFC3302102), the China Postdoctoral Science Foundation (No. 2024M753481), and the Postdoctoral Fellowship Program of China Postdoctoral Science Foundation (No. GZC20232969).

\bigskip

\bibliography{aaai25}

\end{document}